\documentclass[journal]{IEEEtran}
\usepackage{graphicx}
\usepackage{subfigure}
\usepackage{mathrsfs}
\usepackage{amsmath}
\usepackage{multirow}
\usepackage{siunitx}
\usepackage{float}
\usepackage{colortbl}
\usepackage{xcolor}
\usepackage{soul}
\usepackage{bm}
\usepackage{cite}
\usepackage{balance}
\usepackage{stfloats}
\usepackage{algorithm}
\usepackage{algorithmicx}
\usepackage{algpseudocode}
\usepackage{epsfig}
\usepackage{amsfonts,amssymb}
\usepackage{hyperref}
\hypersetup{
	colorlinks=true,
	linkcolor=cyan,
	urlcolor=blue,
	citecolor=cyan,
}

\definecolor{mygray}{gray}{.9}

\begin{document}

\title{Physical Knowledge Enhanced Deep Neural Network for Sea Surface Temperature Prediction}

\author{Yuxin Meng,
        Feng Gao, \emph{Member}, \emph{IEEE},
        Eric Rigall,
        Ran Dong, \\
        Junyu Dong, \emph{Member}, \emph{IEEE},
        Qian Du, \emph{Fellow}, \emph{IEEE}
\thanks{
This work was supported in part by the National Key Research and Development Program of China under Grant 2018AAA0100602, in part by the National Natural Science Foundation of China under Grant 41927805, and in part by the Fundamental Research Funds for the Central Universities under Grant 202113041.
\emph{(Corresponding author: Feng Gao, Junyu Dong)}} %
\thanks{Y. Meng, F. Gao, E. Rigall, R. Dong, and J. Dong are with Institute of Advanced Oceanography, Ocean University of China, Qingdao 266100, China.} 
\thanks{Q. Du is with the Department of Electrical and Computer Engineering, Mississippi State University, Starkville, MS 39762 USA.}
}

\markboth{IEEE Transactions on Geoscience and Remote Sensing}%
{Shell}

\maketitle

\begin{abstract}

Traditionally, numerical models have been deployed in oceanography studies to simulate ocean dynamics by representing physical equations. However, many factors pertaining to ocean dynamics seem to be ill-defined. We argue that transferring physical knowledge from observed data could further improve the accuracy of numerical models when predicting Sea Surface Temperature (SST). Recently, the advances in earth observation technologies have yielded a monumental growth of data. Consequently, it is imperative to explore ways in which to improve and supplement numerical models utilizing the ever-increasing amounts of historical observational data. To this end, we introduce a method for SST prediction that transfers physical knowledge from historical observations to numerical models. Specifically, we use a combination of an encoder and a generative adversarial network (GAN) to capture physical knowledge from the observed data. The numerical model data is then fed into the pre-trained model to generate physics-enhanced data, which can then be used for SST prediction. Experimental results demonstrate that the proposed method considerably enhances SST prediction performance when compared to several state-of-the-art baselines.

\end{abstract}


\IEEEpeerreviewmaketitle

\section{Introduction}\label{S1}

\IEEEPARstart{N}{umerical} models have been a traditional mathematical computation method for prediction of ocean dynamics. According to the statistics from the World Climate Research Program (WCRP), the research community has developed more than 40 ocean numerical models, each of which has its own advantages and characteristics. For instance, the regional ocean model system (ROMS) \cite{roms} has a powerful ecological adjoint module, the fast ocean atmosphere model (FOAM) \cite{foam} is highly effective in global coupled ocean-atmosphere studies, the finite-volume coastal ocean model (FVCOM) \cite{fvcom} is capable of accurately fitting the coastline boundary and the submarine topography. The hybrid coordinate ocean model (HYCOM) \cite{hycom} can implement three varieties of self-adaptive coordinates. These numerical models are not interchangeable and their use depends on the specific application. It should be noted that the various processes of ocean dynamics described in numerical models are based on simplified equations and parameters due to our limited understanding of the ocean. The movements and changes in the real ocean are so diverse and complex that identifying the sources of a certain phenomenon becomes a real challenge. Therefore, searching for new relations or knowledge from historical data is of critical importance to improve the performance of numerical models in the study of ocean dynamics. In this paper, we refer to the capacity that can improve the numerical model as physical knowledge. We assume that the historical data may possess physical knowledge hitherto undiscovered.

\begin{figure}[]
\centering
\includegraphics[width=1.0\linewidth]{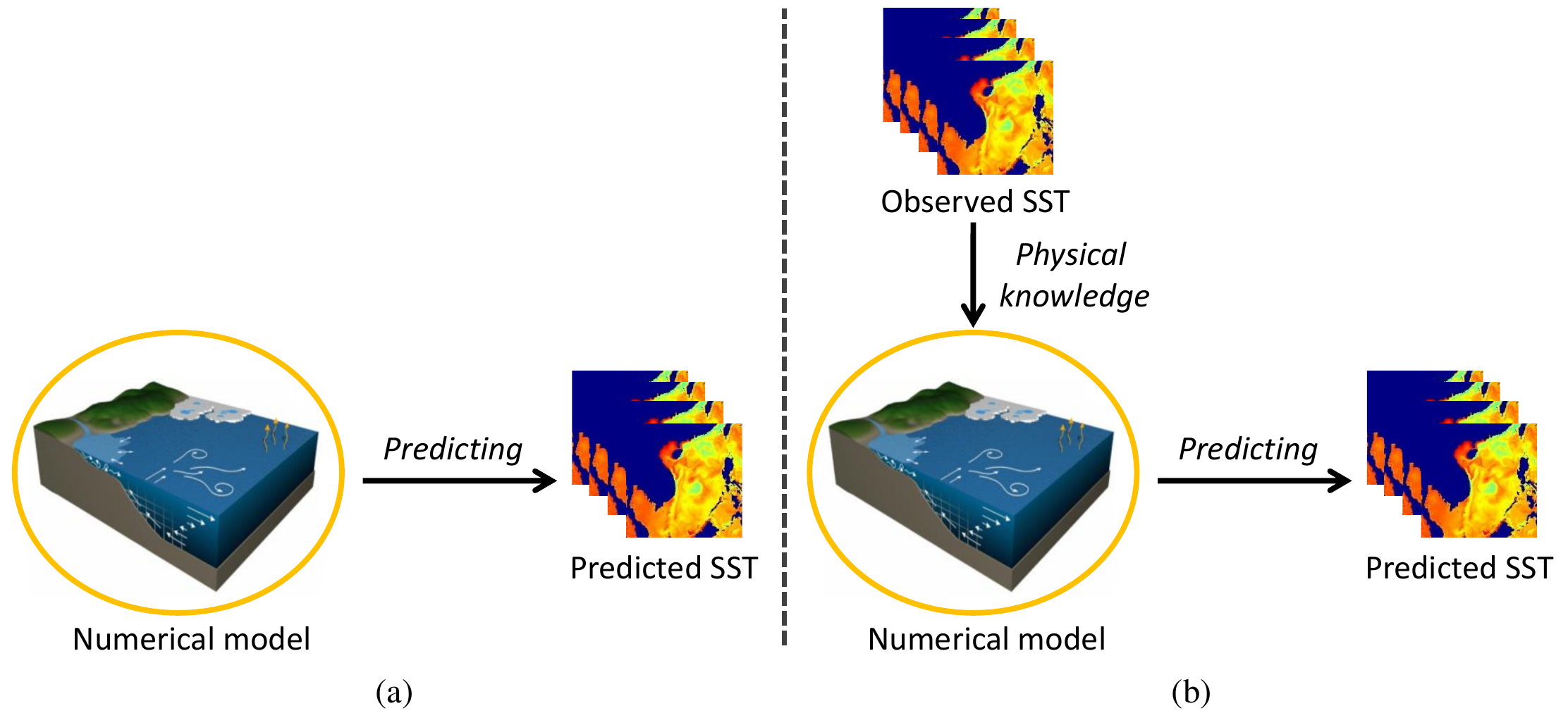} \caption{Conceptual comparison of numerical model and the proposed method on sea surface temperature (SST) prediction. (a) Numerical model. (b) Proposed method for SST prediction. Generative adversarial network is used to transfer the physical knowledge from the historical observed data to the numerical model, and therefore improved the SST prediction performance.}
\label{fig_comp}
\end{figure}

Deep learning has the remarkable ability to learn highly complex functions, transforming the original data into a much higher level of abstraction. In \cite{nature_lecun}, Lecun {\it et al.} described the fundamental principles and the key benefits of deep learning. Recently, deep learning has been applied to a variety of tasks, such as monitoring marine biodiversity \cite{bermant_p, allken_v}, target identification in sonar images \cite{lima_e, xu_l} and sea ice concentration forecasting \cite{sea_ice_ren}. For example, Bermant {\it et al.} \cite{bermant_p} employed convolutional neural networks (CNNs) to classify spectrograms generated from sperm whale acoustic data. Allken {\it et al.} \cite{allken_v} developed a CNN model for fish species classification, leveraging synthetic data for training data augmentation. Lima {\it et al.} \cite{lima_e} proposed a deep transfer learning method for automatic ocean front recognition, extracting knowledge from deep CNN models trained on historical data. Xu {\it et al.} \cite{xu_l} presented an approach combining deep generation networks and transfer learning for sonar shipwrecks detection. Ren {\it et al.} \cite{sea_ice_ren} proposed an encoder–decoder framework with fully convolutional networks that can predict sea ice concentration one-week in advance with high accuracy. Through the application of deep learning-based methods to ocean research, significant improvements have been achieved in terms of classification and prediction performance.

Due to the incomplete physical knowledge in numerical models and the weak generalization performance of neural networks, there aresome efforts to improve prediction performance by combining the advantages of numerical model and nerual networks. In geographical science, this can be achieved in three different ways \cite{reichstein_nature}: \emph{1) Learning the parameters of the numerical model through neural networks.} Neural networks can optimally describe the observed scene from the detailed high-resolution model, but many parameters are difficult to deduce, making their estimation challenging. Brenowitz {\it et al.} \cite{brenowitz} trained a deep neural network based on unified physics parameterization and explained the influence of radiation and cumulus convection. \emph{2) Replacing the numerical model with a neural network.} In this way, the deep neural network architecture can capture the specified physical consistency. Pannekoucke {\it et al.} \cite{pannekoucke} translated physical equations into neural network architectures using a plug-and-play tool. \emph{3) Analyzing the output mismatch between the numerical model and observation data.} Neural networks can be used to identify, visualize, and understand the patterns of the model inaccuracies, and dynamically correct the deviation of the model. Patil {\it et al.} \cite{patil_k} applied the discrepancy between the results of the numerical model and the observational data to train a neural network to predict the sea surface temperature (SST). Ham {\it et al.}  \cite{ham_nature} trained a convolutional neural network based on transfer learning. They first train their model on the numerical model data, and then using reanalysis data to calibrate the model. However, the third approach has been found to suffer from a long-term bias problem, where the prediction performance deteriorates as the prediction days increase.

To address the above issues, in this study, we use the generative adversarial networks (GANs) to transfer the physical knowledge from the historical observed data to the numerical model data, as illustrated in Fig. \ref{fig_comp}. Different from traditional numerical model, the proposed method can correct the physical part in the numerical model data to improve the prediction performance. To be specific, as illustrated in Fig. \ref{fig_framework}, we first acquired the physical feature from the observed data by using a prior network model composed of an encoder and GAN. Thereafter, we obtained the physics-enhanced SST by feeding the numerical model data to the pretrained model. Following that, the physics-enhanced SST were adopted to train a spatial-temporal model for predicting SST. Meanwhile, we performed ablation experiments to take full advantage of the new generated data. 

The main contributions of this paper are threefold:

\begin{itemize} 

\item  To the best of our knowledge, we are the first to transfer physical knowledge from the historical observed data to the numerical model data by using GANs for SST prediction.

\item The difference between the enhanced data based on physical knowledge and the predicted results were exploited to adjust the weight of the model during training.

\item The experimental results indicate that our
proposed method can cover the shortage of physical knowledge in the numerical model and improve the prediction accuracy.

\end{itemize}

The rest of the paper is organized as follows. Section II introduces the literature review related to our method, while our method design is detailed in Section III. Then the experimental results are shown in Section IV. Section V finally concludes this paper.

\section{Background}\label{S2}

\subsection{Generative Adversarial Network}\label{S2.1}

In 2014, Goodfellow {\it et al.} \cite{goodfellow14_nips} put forward a novel framework of generative model trained on an adversarial manner. In their method, a generative model G and a discriminative model D were trained simultaneously. 
The model G was applied to indirectly capture the distribution of the input data through model D and generate similar data. 
While model D estimates the probabilities that its input samples came from training data instead of model G. The training process of G was driven by the probability errors of D. In this adversarial process, G and D guide the learning and gradually strengthen each other's ability to achieve outstanding performance. 

GANs have been applied in physical-relevant tasks. For example, Yang {\it et al.} \cite{PI_SDE} applied physics-informed GANs to deal with high-dimensional problems and solved stochastic differential equations, L{\"u}tjens {\it et al.} \cite{PI_CFV} produced more realistic coastal flood data by using GANs to learn the features in the numerical model data, Zheng {\it et al.} \cite{PI_SI} inferred the unknown spatial data with the potential physical law that is learned by GANs. However, these works performed their model by using GAN to replace the entire numerical model, which is quite different from our work. In this paper, we adopt GAN model to transfer the physical knowledge from the observed data to the numerical model data, in order to correct and improve the physical feature in the numerical model. In addition, existing methods only learn a deterministic model without considering whether the code generated by the encoder is in accordance with the semantic knowledge learned by the GAN. 

\subsection{Convolutional Long Short-Term Memory}\label{S2.2}

In 2015, ConvLSTM \cite{shix} was proposed to solve the precipitation nowcasting. The network structure of ConvLSTM is able to capture local spatial features as in classical convolutional neural networks (CNN) \cite{guj} while building a sequential relationship, inherited from Long Short-Term Memory (LSTM) blocks. Moreover, the authors conducted experiments to show that ConvLSTM is able to perform better than LSTM on spatial-temporal relationship. 
Apart from weather prediction tasks, ConvLSTM can be applied to various spatial-temporal sequential prediction problems, for example, action recognition \cite{geh,chew}.

\begin{figure*}[ht]
\vspace*{-4mm}
\begin{center}
\includegraphics[width=1.0\linewidth,angle=0]{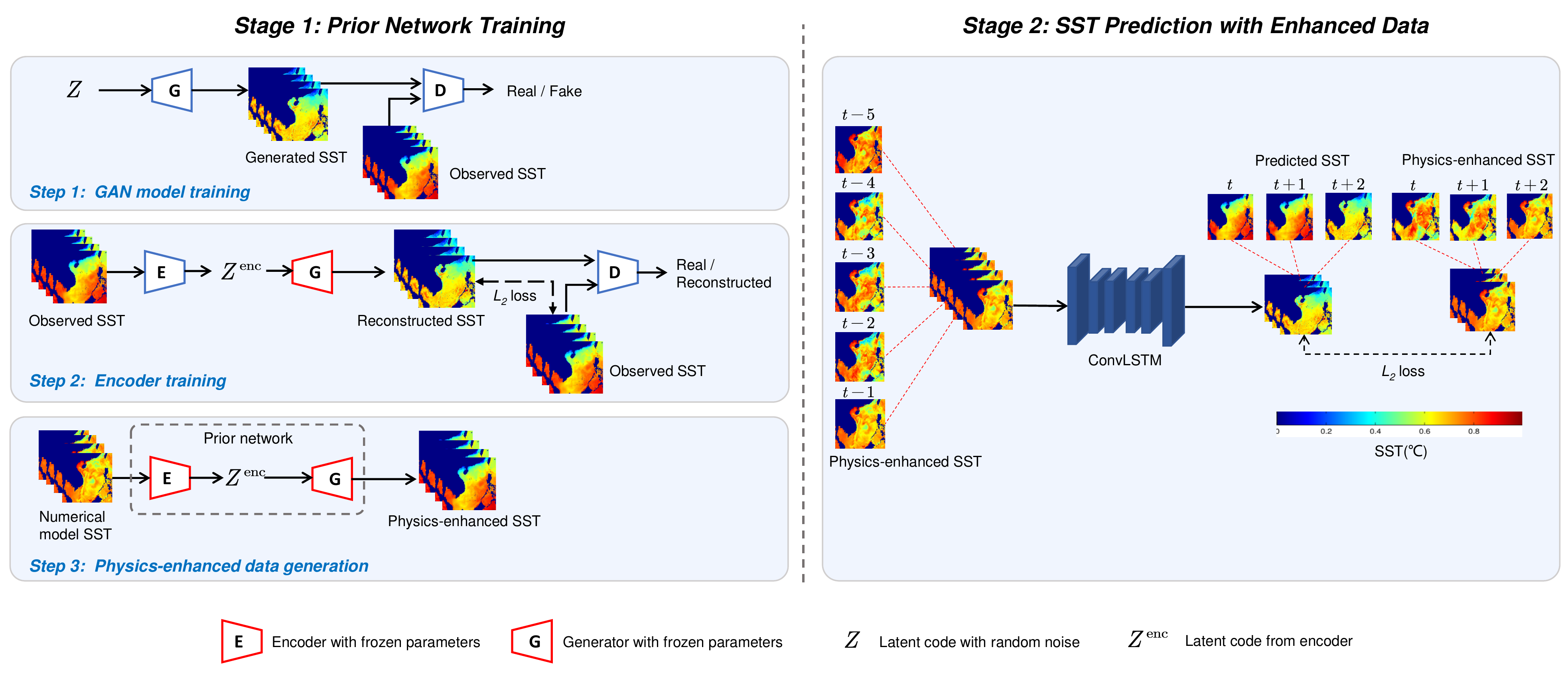} 

\end{center}
\vspace*{-4mm}
\caption{Illustration of the proposed SST prediction method. It consists of two stages: Prior network training and SST prediction with enhanced data. In the first stage, a prior network is trained to generate physics-enhanced SST. In the second stage, the physics-enhanced SST are used for SST prediction via ConvLSTM.}
\label{fig_framework}
\vspace*{-1mm}
\end{figure*}

\subsection{Sea Surface Temperature Prediction}\label{S2.3}

 Lins \emph{et al.} \cite{lins} investigated SST in tropical Atlantic using an SVM. Patil {\it et al.} \cite{patil_kb} adopted an artificial neural network to predict the sea surface temperature. It performs well only in the case of forecasting with the lead time from 1 to 5 days and then the accuracy declined. Zhang \emph{et al.} \cite{zhangq} applied LSTM to predict SST. Yang \emph{et al.} \cite{yangy} predicted SST by building a fully connected LSTM model. From another perspective, Patil \emph{et al.} \cite{patil} used a wavelet neural network to predict daily SST, while Quala \emph{et al.} \cite{quala} proposed patch-level neural network method for SST prediction. However, these methods only rely on data and ignore the physical knowledge behind them. Ham {\it et al.}  \cite{ham_nature} adopted transfer learning to predict ENSO and classify them. In this work, we conduct comparative experiments and the results point out that our method reduces the short-term errors as well as the long-term bias. 

\subsection{Data Augmentation}\label{S2.4}
Shorten \emph{et al.} \cite{shorten_c} reviewed recent techniques of image data augmentation for deep learning. The purpose of data augmentation is to enhance the representation capability of neural networks and learn the distribution of original data better. In recent years,  two kinds of data augmentation techniques have been commonly used: data transformation and resampling. The data transformation approach includes geometric transformation  \cite{hessam_b}, color space transformation \cite{ken_c,aramzazu_j,quanzeng_y}, random erasing \cite{zhun_z,terrance_v,agnieszka_m}, adversarial training \cite{seyed-mohsen,jiawei_s,michal_z,shuangtao_l} and style transfer \cite{leon_ag,dmitry_u,philip_tj,josh_t}. The resampling technique lays particular emphasis on new instance composition, such as image mixup \cite{cecilia_s,daojun_l,ryo_t}, feature space enhancement \cite{tomohiko_k,terrancev} and generative adversarial network (GAN) \cite{goodfellow14_nips}. Geometric transformation can acquire nice performance,
such as image flip, crop, rotation, translation, and noise injection \cite{francisco_jm-b}. 
The experimental results in \cite{shorten_c} showed that the random cropping technique performed well. Color space transformation suffers from a large memory consumption and long computing time. Random erasing techniques can improve the network robustness in occlusion cases by using masks. Although adversarial training can also improve robustness, the finite number of natural adversarial samples largely limits the network performance in practice. The neural style transfer approach is only effective for specific tasks, while its practical application is limited. The feature space augmentation implements the capability of interpolating representations in the feature space. 
GAN-based augmentation techniques have been applied to achieve current state-of-the-art network performance\cite{maayan_f-a}. However, there does not exist an effective data augmentation method that could exploit the merits of the numerical model and deep learning. In this paper, we aim to propose a novel data enhancement technique based on physical knowledge. The proposed technique achieves better performance than GAN-based augmentation. 

\section{Proposed Method}\label{S3}

Numerical model can predict the spatial distribution of SST and its global teleconnections together. It performs well at short-leads for SST prediction. Nevertheless, we argue that transfer the physical knowledge from the observed data can further improve the performance of numerical model for SST prediction. To this end, we adopt GANs to learn the physical knowledge in the observed data. 

Zhu {\it et al.} \cite{zhuj} proposed a GAN inversion method that not only faithfully reconstructs the input data, but also ensures that the inverted latent code is semantically meaningful. They demonstrated that learning the pixel values of the target image alone is insufficient, and that the learned features are unable to represent the image at the semantic level. Inspired by this work, we design an encoder in GAN to learn physical knowledge from the observed data, referred to as the prior network. This prior network not only learns the pixel values of the target observed data, but also captures the physical information. It effectively improves the SST prediction accuracy.

Next, we present the proposed method as follows: 1) Overview of the method, 2) Prior network, 3) SST prediction with enhanced data.

\subsection{Overview of the Method}

In this subsection, we summarize the proposed SST prediction method and describe the input and output of each stage in detail. As illustrated in Fig. \ref{fig_framework}, the proposed SST prediction method consists of two stages: Prior network training and SST prediction with enhanced data. 

\textbf{1) Prior network training.} This stage consists of three steps. In the first step, the observed SST (GHRST data) is used for GAN model training. In the second step, the pretrained generator and GHRSST data are used to train the encoder. In the third step, the pretrained generator and encoder are combined into the prior network. The prior network is used to transfer the physical knowledge from the observed data to the numerical model. The numerical model SST (HYCOM data) is then fed into the prior network to enhance its feature representations.

\textbf{2) SST prediction with enhanced data.} The physics-enhanced data are fed into ConvLSTM model for SST prediction. The SST of the next day, next 3 days, and next 7 days are predicted separately.

It should be noted that most existing works \cite{zhangq} \cite{yangy} only use the observed data for ConvLSTM training. By contrast, our method takes advantage of physics-enhanced data for ConvLSTM training. Next, we describe prior network training and SST prediction with enhanced data in details.

\subsection{Stage 1: Prior Network Training}

We construct a prior network to learn the physical knowledge in the observed data and keep its semantic/physical information constant after training. As illustrated in Fig. \ref{fig_framework}, the prior network training is comprised of three steps: GAN model training, encoder training, and physics-enhanced data generation. Next we provide detailed descriptions of each step.

\textbf{GAN Model Training.} The GAN model is used to learning the data distribution from the observed SST. The objective function is as follows:

\begin{align}\label{eq1}
\mathop{\min}\limits_{G}\mathop{\max}\limits_{D}L(D, G)
=&\mathop{\mathbb{E}}\limits_{m\sim p_{data}(m)}[\log D(m)]  \nonumber \\
&+\mathop{\mathbb{E}}\limits_{z\sim p_{z}(z)}[\log(1-D(G(z)))],
\end{align}
where $z$ represents the random vector fed into the generator $G$, and $m$ refers to the observed SST data. The generator $G$ aims to capture the data distribution from the observed SST data. The discriminator $D$ aims to estimate the probability that whether the input sample comes from real data or from the generator $G$. Here $p_{data}$ denotes to the observed data distribution, and $p_z$ denotes the noise distribution. Through adversarial training, GAN captures the physical information of the observed SST data, and hence well-trained generator $G$ can produce high quality SST data.

The training process of the GAN model is summarized in Algorithm~\ref{ALG1}. We train the model over the observed SST until the generator $G$ captures the physical features from the observed SST data.

\begin{algorithm}[bp!]
\caption{: GAN Model Training}
\label{ALG1}
\begin{algorithmic}[1]
\Require
Random noise vector $z$, observed training data $m$, numbers of epochs $n_{1}$ 
\Ensure
Generator $G$ and discriminator $D$\
\While {not converged}
  \For{$t=0,\cdots,n_{1}$}
    \State Sample image pair $\{z^{i}\}_{i=1}^{N}$ and $\{m^{i}\}_{i=1}^{N}$;
    \State Update $D$ by gradient descend based on Eq. \ref{eq1};
    \State Update $G$ by gradient descend based on Eq.  \ref{eq1};
  \EndFor
\EndWhile
\end{algorithmic}
\end{algorithm}

\textbf{Encoder Training.} The observed SST data are fed into an encoder $E$ to generate latent code $Z^\text{enc}$. Through adversarial training, the encoder $E$ captures the semantic/physical information from the observed SST. It should be noted that the parameters of the generator $G$ are fixed, and the discriminator $D$ aims to identify the real samples from the generated samples. $D$ and $E$ are trained as follows:

\begin{align}\label{eq2}
\mathop{\min}\limits_{\varTheta_{E}}L_{E} =& ||m-G(E(m))||_{2} \nonumber \\ &-\lambda_{adv}\mathop{\mathbb{E}}\limits_{m\sim p_{data}}[D(G(E(m)))]  \nonumber \\
& +\lambda_{vgg}||F(m)-F(G(E(m)))||_{2},
\end{align}

\begin{align}\label{eq3}
\mathop{\min}\limits_{\varTheta_{D}}L_{D} =&\mathop{\mathbb{E}}\limits_{m\sim p_{data}}[D(G(E(m)))] \nonumber \\
&-\mathop{\mathbb{E}}\limits_{m\sim p_{data}}[D(m)]   \nonumber \\
& +\frac{\gamma}{2}\mathop{\mathbb{E}}\limits_{m\sim p_{data}}[||\nabla_{m}D(m)||_{2}^{2}],
\end{align}
where $F(\cdot)$ represents feature extraction via the VGG network. VGG network stands for the network proposed by Visual Geometry Group \cite{vgg}, and it is a classical deep convolutional neural network.

The encoder training is described in Algorithm \ref{ALG2}. The parameters of the generator $G$ are fixed, while the parameters of encoder $E$ and discriminator $D$ are updated based on Eq. \ref{eq2} and Eq. \ref{eq3}, respectively. 

\begin{algorithm}[tp!]
\caption{: Encoder Training}
\label{ALG2}
\begin{algorithmic}[1]
\Require
Observed training data $m$, $\lambda_{adv}$, $\lambda_{vgg}$, $\gamma$, number of epochs $n_{2}$
\Ensure
Encoder $E$ and discriminator $D$
\While {not converged}
  \For{$t=0,\cdots,n_2$}
    \State Sample images $\{m^{i}\}_{i=1}^{N}$;
    \State Update $D$ by gradient descend based on Eq. \ref{eq3};
     \State Update $E$ by gradient descend based on Eq. \ref{eq2};  
  \EndFor
\EndWhile
\end{algorithmic}
\end{algorithm}

\textbf{Physics-Enhanced Data Generation.} The well-trained encoder $E$ and generator $G$ are combined into the prior network. Numerical model data are fed into the prior network to generate physics-enhanced data, in which the incorrect components would be restored. 

The motivation of Stage 1 is to construct a prior network which can rectify the incorrect components in the numerical model data. To this end, we firstly devise a GAN model which captures the data distribution from the observed SST and can generate high-quality SST data. Subsequently, the encoder is trained to guarantee that the generated latent codes preserve the semantic/physical information in the observed SST. We argue that through adversarial learning, the prior network (consisting of the encoder and generator) can rectify the incorrect parts in the input data, since the physical knowledge has been embedded in the prior network. Consequently, in the third step, when the numerical model data are fed into the prior network, the embedded physical knowledge can correct the incorrect components in the numerical model data.

\subsection{Stage~2:~SST Prediction with Enhanced Data}

ConvLSTM is an effective tool for predicting spatial-temporal data. It is a recurrent neural network that incorporates convolutional blocks in both the input-to-state and state-to-state transitions. Unlike the traditional LSTM layer, ConvLSTM not only preserves the sequential relationship but also extracts spatial features from the data. In this way, we can leverage it to capture robust spatial-temporal features. The objective function of ConvLSTM is formulated as follows:

\begin{align}\label{eq4}
i_{t}=&\sigma(W_{xi}\ast X_{t}+W_{hi}\ast H_{t-1}+W_{ci}\circ C_{t-1}+b_{i}) \nonumber \\
f_{t}=&\sigma(W_{xf}\ast X_{t}+W_{hf}\ast H_{t-1}+W_{cf}\circ C_{t-1}+b_{f}) \nonumber \\
C_{t}=&f_{t}\circ C_{t-1}+i_{t}\circ tanh(W_{xc}\ast X_{t}+W_{hc}\ast H_{t-1}+b_{c})\nonumber \\
O_{t}=&\sigma(W_{xo}\ast X_{t}+W_{ho}\ast H_{t-1}+W_{co}\circ C_{t}+b_{o})\nonumber \\
H_{t}=&O_{t}\circ tanh(C_{t}),
\end{align}
where $\ast$ denotes the convolution operation, $\circ$ denotes the Hadamard product, $W$ and $b$ are the corresponding weights and bias respectively, $H_{t-1}$ and $X_{t}$ are previous output and current input respectively. The input gate $i_{t}$, forget gate $f_{t}$ and output gate $O_{t}$ aim to protect and control the cell state $C_{t}$. The three-dimensional tensors $X_{t}$, $i_{t}$, $f_{t}$, $C_{t}$, $O_{t}$ and the two-dimensional matrix $H_{t}$ refer to the spatial information. 

The physics-enhanced SST data are fed into the ConvLSTM model for SST prediction as follows:

\begin{equation}\label{eq5}
  L_{CL} = \lVert \text{ConvLSTM}(Ph_{i-t},...,Ph_{i-1})-Ph_{i} \rVert_2,
\end{equation}
where $CL$ refers to the ConvLSTM model. The physics-enhanced data are marked as $Ph$. The past $t$ days' data from physics-enhanced data are fed into the ConvLSTM and the output is compared with $Ph_{i}$. Here $t$ denotes the number of past days used for prediction. It is a critical parameter that may affect the SST prediction performance. Comprehensive analysis of $t$ can be found in Section IV. B.

\begin{algorithm}[tp!]
\caption{: SST Prediction with Enhanced Data}
\label{ALG3}
\begin{algorithmic}[1]
\Require
Physical numerical model training data $Ph$, number of epochs $n_{3}$
\Ensure
Convolutional LSTM $ConvLSTM$ 
\While {not converged}
  \For{$t=0,\cdots,n_3$}
    \State Sequence sample image pair $\{Ph^{i}\}_{i=1}^{N}$;
     \State Update $ConvLSTM$ by gradient descend based on Eq. \ref{eq5}; 
  \EndFor
\EndWhile
\end{algorithmic}
\end{algorithm}

The weights obtained by the generator are reused in Algorithm~\ref{ALG2}, where only the generator weights are fixed. The introduced encoder and the discriminator go through another training process over the observed SST. Their weights are updated based on Eq. \ref{eq2} and Eq. \ref{eq3}, respectively. After training, the code generated by the encoder would embody the learned physical knowledge. 

Finally, we acquire the data reinforced based on physical knowledge using the above pre-trained model. The weights of the generator and the encoder from Algorithm~\ref{ALG2} are reused and the numerical model SST is exploited to produce physics-reinforced numerical model data. 

In Algorithm~\ref{ALG3}, the physical knowledge-enhanced data are leveraged to train a spatial-temporal ConvLSTM model for SST prediction. In this paper, the SST of the next day, the next 3 days and the next 7 days are predicted separately. For this part, we conducted an ablation study to make use of the reinforced data effectively.

\section{Experimental Results and Analysis}\label{S4}

\subsection{Study Area and Experiment Settings}

The South China Sea is located in the Western Pacific Ocean, in southern mainland China. Its area is about 3.5 million square kilometers with an average depth of 1, 212 meters. In this paper, the selected study area is (\ang{3.99}N$\sim$\ang{24.78}N, \ang{98.4}E$\sim$\ang{124.4}E).

We use the high resolution satellite remote sensing data from GHRSST (Group for High Resolution Sea Surface Temperature) \cite{ghrsst} as the observed data. GHRSST provides a variety of sea surface temperature data, including satellite swath coordinates, gridded data, and gap-free gridded products. Herein, we have employed gap-free gridded products, which are generated by combining complementary satellite and in situ observations within an Optimal Interpolation framework. The HYCOM \cite{hycom_data} is selected as the numerical model. Their spatial resolutions are 1/\ang{20}$\times$1/\ang{20} and 1/\ang{12}$\times$1/\ang{12}, respectively. The temporal resolution is one day. The data from May 2007 to December 2013 are used for training, while the remaining data from January 2014 to December 2014 are used for testing. It should be noted that we use cloudless data provided by GHRSST. The data were captured by microwave instruments which can penetrate through clouds. Hence, the data have full coverage of the study area. In addition, the accurate time of every pixel in GHRSST SST product is the same.

The Z-score standardization was utilized for preprocessing as:
\begin{equation}\label{eq6}
  z = \frac{x-\mu}{\sigma},
\end{equation}
where $x$ denotes the GHRSST and HYCOM model SST, $z$ denotes the normalized data, $\mu$ and $\sigma$ denote the mean value and standard deviation respectively. We converted the data into $256 \times 256$ square-shaped heat maps.

More specifically, the GHRSST data and 512-dimensional random vector are utilized in the first step of prior network training. The size of the input GHRSST data is $N \times H \times W$, where $N$ represents the batch size, $H$ indicates the height of the input data, and $W$ denotes the width of the input data. For the second stage of prior network, we only employ GHRSST data for encoder training. The sizes of inputs and outputs for both stages are $N \times H \times W$. Similarly, in the third step of prior network training, the HYCOM SST data is fed into the pretrained model. Here, the sizes of both the inputs and the outputs are $N \times H \times W$. In our implementations, we set $N$ to 2430, while $H$ and $W$ are both set to 256.

\begin{figure}[ht]
\vspace*{-1mm}
\begin{center}
\rotatebox{0}{\includegraphics[width=0.98\columnwidth,angle=0]{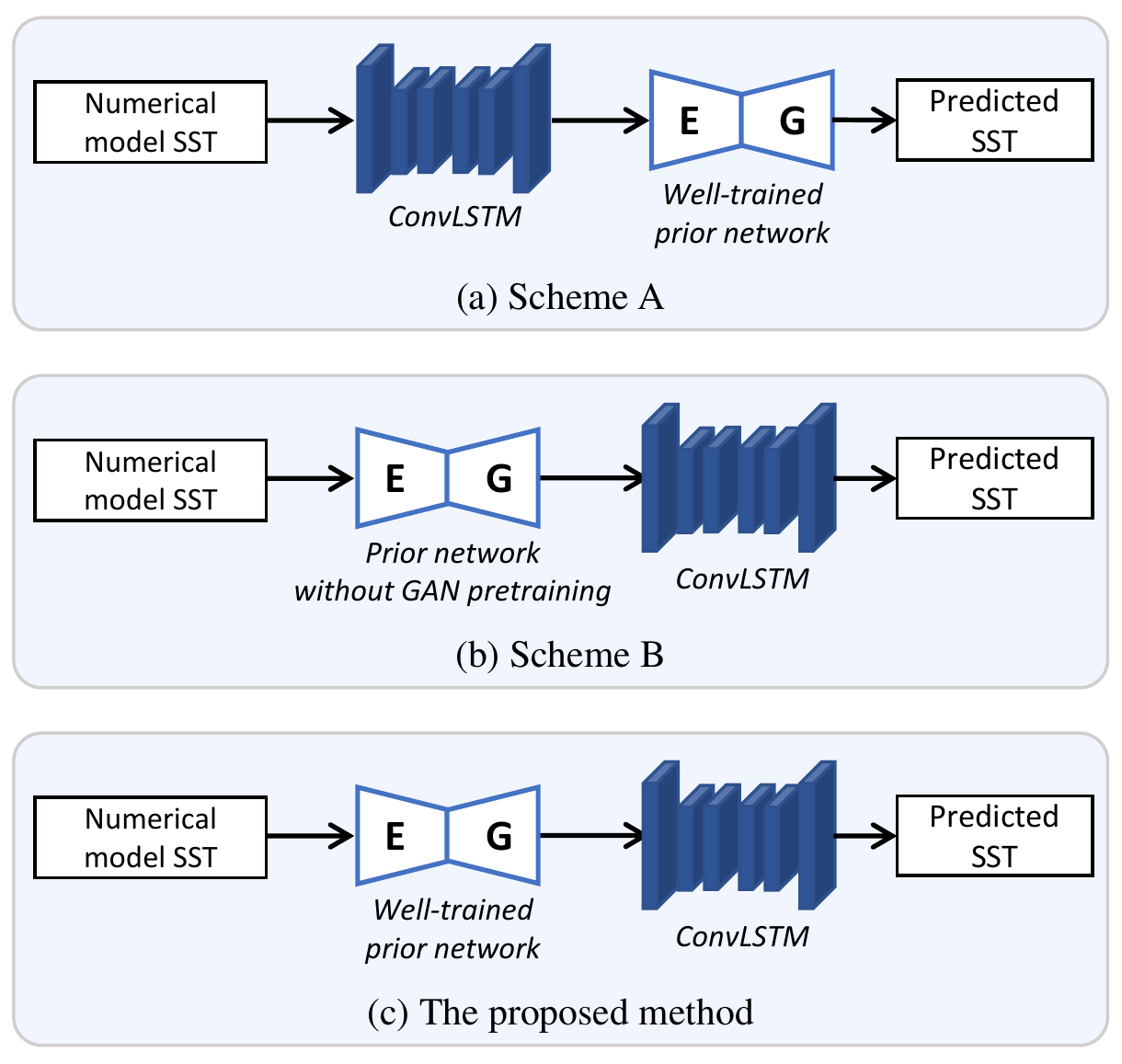}}
\end{center}
\caption{Illustration of three models used in ablation study. (a) Scheme A: The numerical model SST data are first fed into the ConvLSTM, and then the output are fed into the well-trained prior network. The sequence of prior network and ConvLSTM is replaced. (b) Scheme B: The prior network has not been well-trained. Specifically, GAN model training in prior network has been omitted. (c) The proposed method.}
\label{fig_ablation}
\vspace*{-1mm}
\end{figure}

We conducted extensive experiments on an NVIDIA GeForce 2080Ti with 8 GPUs. The prior network uses the same network structure and configuration as mentioned in \cite{zhuj} to  acquire the physical knowledge from the historical observed data. Then the obtained physical knowledge is transferred to the numerical model data for the sake of restoring and improving the incorrect components in the numerical model. The configuration for the ConvLSTM model used in this paper is the same as the ConvLSTM model in Shi's work \cite{shix}. The GHRSST SST dataset is utilized as the benchmark for comparison and assessment in this paper.

\subsection{Influences of the Past Day Number for SST Prediction}

As mentioned in Section III. C, $t$ denotes the number of past days used for prediction. It is a critical parameter that may affect the SST prediction performance. In this paper, we attempt to predict the next one-day, three-day and seven-day's SST. We implemented extensive experiments to find the proper number of past days for the future SST prediction. The Root Mean Square Error (RMSE) and the coefficient of determination ($R^2$) are applied as the evaluation criteria. Lower RMSE and higher $R^2$ values indicate more accurate results. 

Table~\ref{table_one_day} lists the prediction results for the next day by using the past one day, three days and five days' data, separately. It can be observed that the proposed model performs best when using the past five days' data, where the RMSE and the $R^2$ results are 0.3618 and 0.9967 respectively. They are slightly better than the other schemes. Compared to the other two schemes, the RMSE and $R^2$ values improve by 0.0086, 0.001 and 0.0028, 0.0006. Hence, the past five days' data are adopted for the next one-day SST prediction. 

\begin{table}[tp!]
\caption{Experimental Results of Different Number of Past Days for the Next One-Day SST Prediction} 
\renewcommand\arraystretch{1.5}
\label{table_one_day} 
\begin{center}
\begin{tabular}{|c||c|c|}
\hline
\multirow{1}{*}{~~~ Days ~~~}  &  {~~~ RMSE ($^{\circ}$C) $\downarrow$ ~~~} &{~~~ $R^{2}$ $\uparrow$ ~~~} \\
\cline{2-2} \hline \hline
1 & 0.3704 & 0.9957\\ 
3 & 0.3646 & 0.9961\\ 
5 & \textbf{0.3618} & \textbf{0.9967}\\ \hline 
\end{tabular}
\end{center}
\end{table}

We analyze the influences of $t$ for the next three-day SST prediction in Table~\ref{table_three_days}. It can be seen that the longer historical data was used, the better prediction performance was achieved. The RMSE value using the past seven days' data  achieves the best performance. It is improved by 0.0025 compared to that using the past five days' data. Meanwhile, the $R^2$ performs the best using the past seven days' data compared to the other two schemes. Therefore, the past seven days' data were used for the next three-day SST prediction.

\begin{table}[tp!]
\caption{Experimental Results of Different Number of Past Days for the Next Three-Day SST Prediction}
\renewcommand\arraystretch{1.5}
\label{table_three_days} 
\begin{center}
\begin{tabular}{|c||c|c|}
\hline
\multirow{1}{*}{~~~ Days ~~~}  &  {~~~ RMSE ($^{\circ}$C) $\downarrow$ ~~~} &{~~~ $R^{2}$ $\uparrow$ ~~~} \\
\cline{2-2} \hline \hline
3 & 0.3939 & 0.9948 \\ 
5 & 0.3681 & 0.9943\\ 
7 & \textbf{0.3656} & \textbf{0.9961}\\ \hline 
\end{tabular}
\end{center}
\end{table}

The experimental results of the next seven-day SST prediction is illustrated in Table \ref{table_seven_days}. As can be seen that the prediction results using the past ten days' data achieves the best performance. Therefore, we exploit the past ten days' data for the next seven-day SST prediction. 

\begin{table}[tp!]
\caption{Experimental Results of Different Number of Past Days for the Next Seven-Day SST Prediction}
\renewcommand\arraystretch{1.5}
\label{table_seven_days} 
\begin{center}
\begin{tabular}{|c||c|c|}
\hline
\multirow{1}{*}{~~~ Days ~~~}  &  {~~~ RMSE ($^{\circ}$C) $\downarrow$ ~~~} &{~~~ $R^{2}$ $\uparrow$ ~~~} \\
\cline{2-2} \hline \hline
7 & 0.3872 & 0.9934 \\ 
9 & 0.3798 & 0.9945\\ 
10 & \textbf{0.3794} & \textbf{0.9961}\\ \hline 
\end{tabular}
\end{center}
\vspace*{-5mm}
\end{table}

\subsection{Ablation Study}

To verify the effectiveness of the prior network and GAN training, we conduct ablation experiments. As illustrated in Fig. \ref{fig_ablation}, two variants are designed for comparison as follows:

\begin{itemize}

\item \textit{Scheme A}. The sequence of prior network and ConvLSTM is replaced. The numerical model SST data are first fed into the ConvLSTM, and then the output are fed into the well-trained prior network.

\item \textit{Scheme B}. The prior network has not been well-trained. Specifically, GAN model training (the first step in Fig. \ref{fig_framework}) in prior network training has been omitted.

\end{itemize}

The experimental results are shown in Table \ref{table_ablation}. As can been seen that our method achieves the best RMSE and $R^2$ values. Specifically, the proposed method outperforms Scheme A, which demonstrates that the correct sequence of prior network and ConvLSTM can boost the SST prediction performance. It is evident that the prior network effectively restores the incorrect components of the numerical model data, and the restored data perform better in SST prediction. Futhermore, the proposed method has superior performance over Scheme B, which demonstrates that the GAN modeling is an essential step. GAN modeling can learn the data distribution of the observed SST, and helps the prior network capture better physical information from the observed SST. To sum up, in the proposed method, we use adversarial learning for prior network pretraining, which can effectively transfer physical knowledge from the observed SST data to the prior network. It can guide fast training convergence, and improve the SST prediction performance.

\begin{table}[tp!]
\vspace*{-1mm}
\caption{Experimental Results (Average$\pm$STD) of Ablation Studies Averaged over 10 Random Runs}
\renewcommand\arraystretch{1.5}
\label{table_ablation} 
\vspace*{-4mm}
\begin{center}
\begin{tabular}{|c||c|c|c|}
\hline
\multirow{2}{*}{Model}  & \multicolumn{3}{c|}{RMSE ($^{\circ}$C) $\downarrow$} \\
\cline{2-4} & ~~~~1 day~~~~ & ~~~~3 days~~~~ & ~~~~7 days~~~~ \\ \hline\hline
Scheme A & 0.9126$\pm$0.0021 & 0.8340$\pm$0.0030 & 0.8806$\pm$0.0078 \\   
Scheme B & 0.7991$\pm$0.0034 & 0.7146$\pm$0.0041 & 0.7394$\pm$0.0433 \\ 
Ours~~~ & \textbf{0.3618$\pm$0.0017}& \textbf{0.3656$\pm$0.0015} &\textbf{ 0.3794$\pm$0.0014} \\ \hline \hline
\multirow{2}{*}{Model}  & \multicolumn{3}{c|}{$R^{2}$ $\uparrow$} \\
\cline{2-4} & ~~~~1 day~~~~ & ~~~~3 days~~~~ & ~~~~7 days~~~~ \\ \hline   \hline
Scheme A & 0.9694$\pm$0.0032 & 0.9768$\pm$0.0014 & 0.9715$\pm$0.0037 \\  
Scheme B & 0.9707$\pm$0.0035 & 0.9771$\pm$0.0026 & 0.9779$\pm$0.0034 \\ 
Ours~~~ & \textbf{0.9967$\pm$0.0012} & \textbf{0.9961$\pm$0.0013} &\textbf{0.9961$\pm$0.0024} \\\hline
\end{tabular}
\end{center}
\vspace*{-5mm}
\end{table}

\subsection{Experimental Results and Discussion}

\begin{figure}
  \centering
  \includegraphics[width=3.0in]{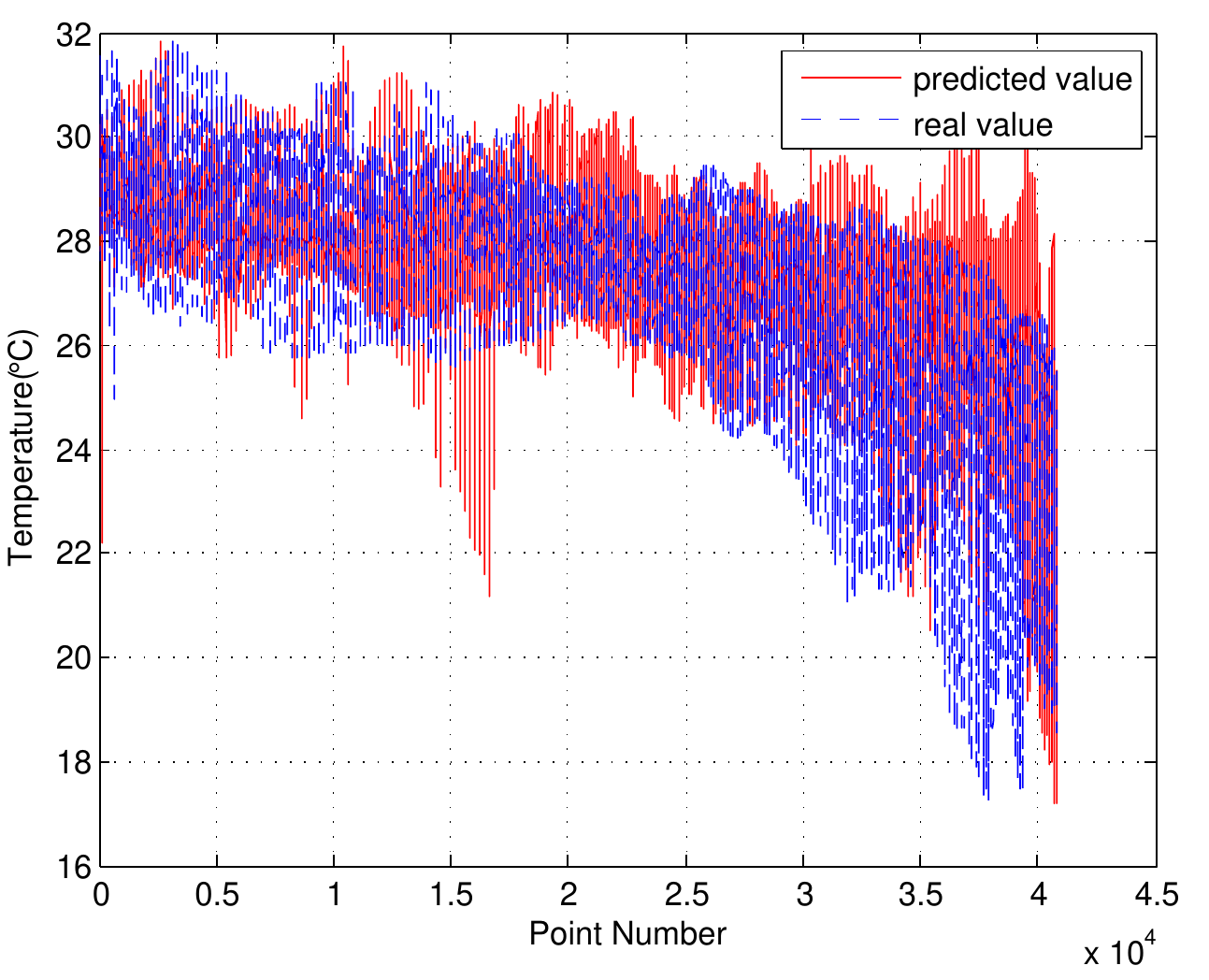}
  \caption{Next one-day SST prediction results versus the observed ground truth data.}
  \label{fig_1_day}
\end{figure}

\begin{figure}
  \centering
  \includegraphics[width=3.0in]{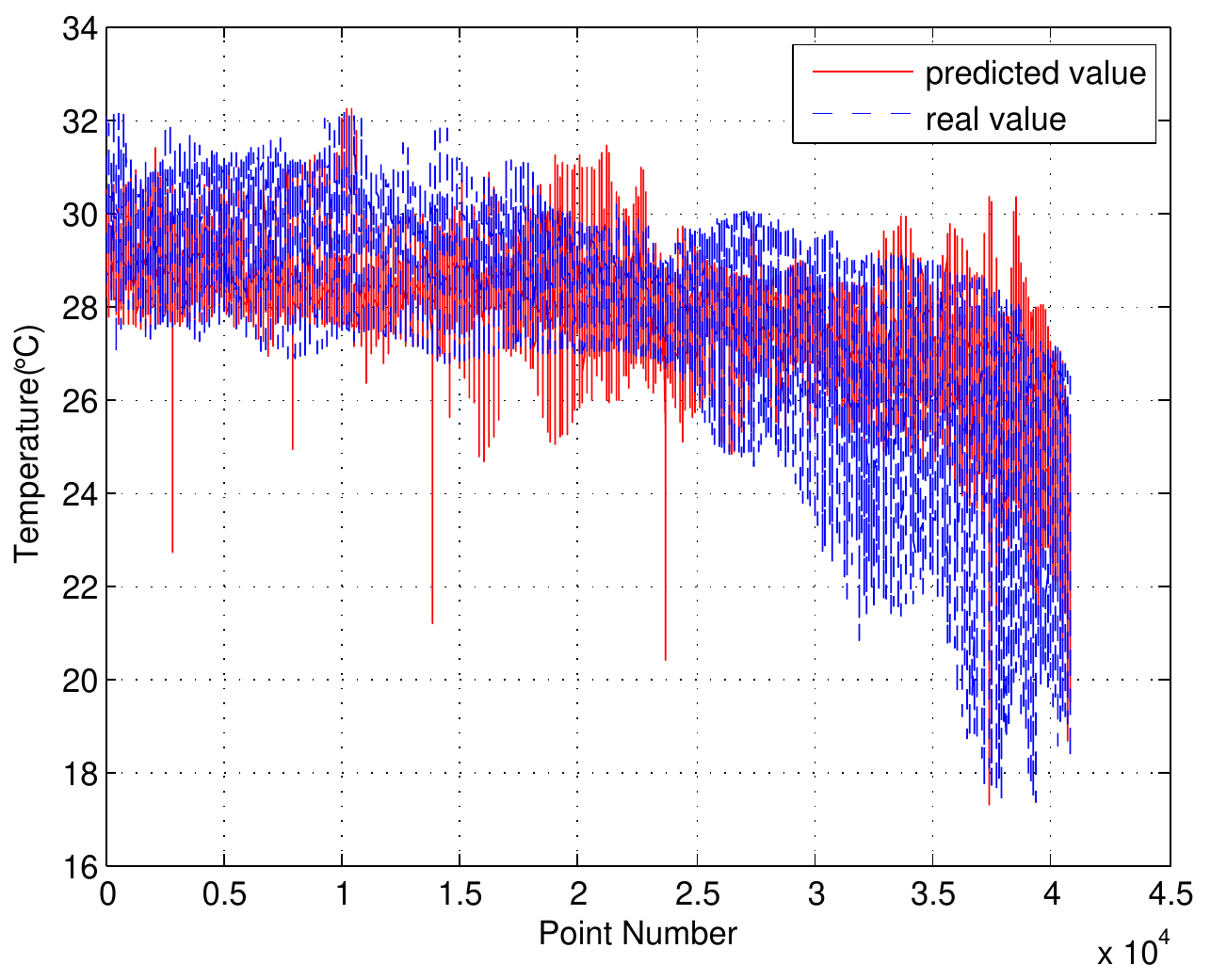}
  \caption{Next three-day SST prediction results versus the observed ground truth data.}
  \label{fig_3_days}
\end{figure}

\begin{figure}
  \centering
  \includegraphics[width=3.0in]{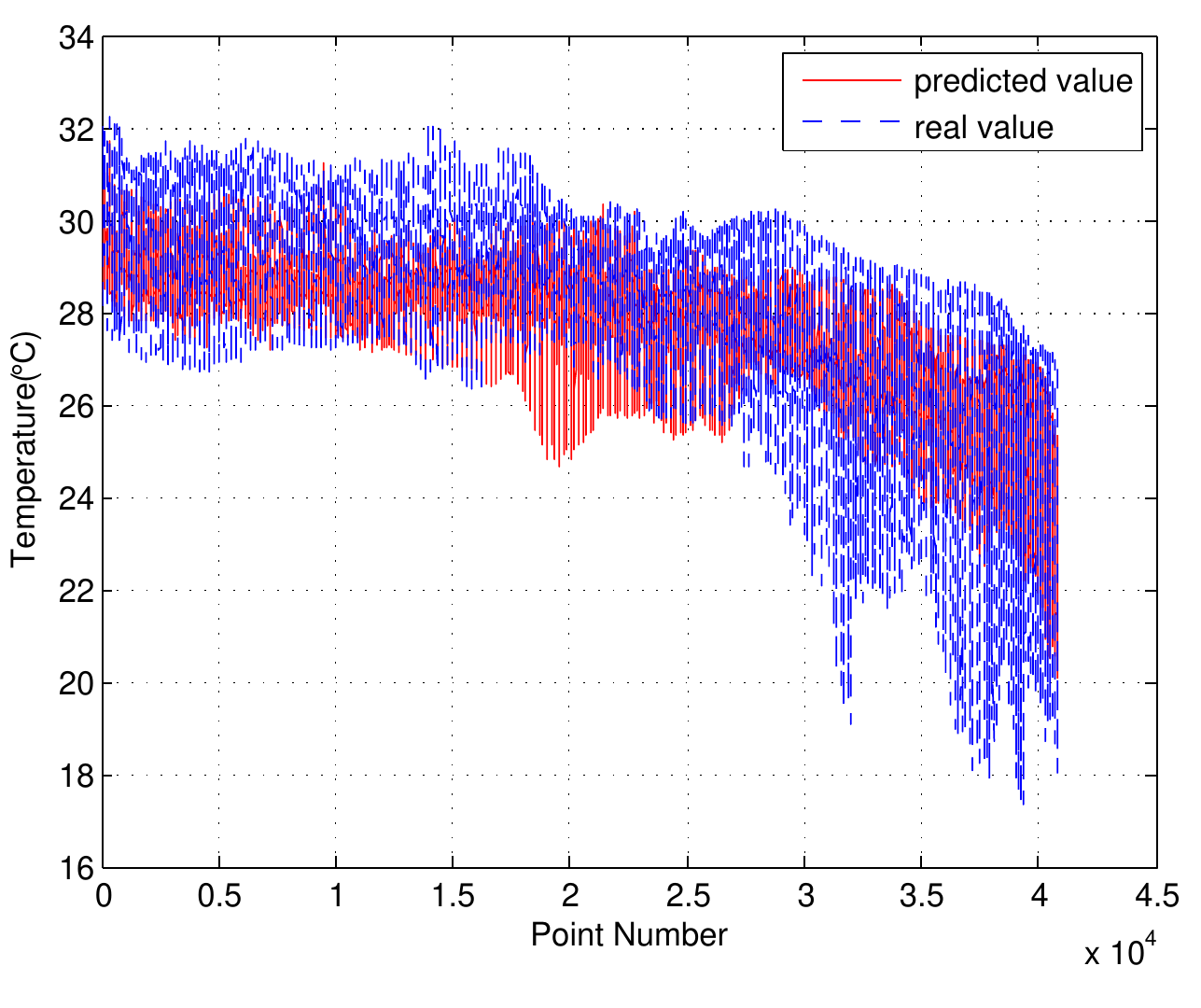}
  \caption{Next seven-day SST prediction results versus the observed ground truth data.}
  \label{fig_7_days}
\end{figure}

Fig. \ref{fig_1_day} compares the predicted next one-day SST with the observed ground truth data. We can see that the predicted results of our method match well with the observed data. Similarly, the observed data and the corresponding predicted SST for the next three days and seven days are displayed in Fig. \ref{fig_3_days} and Fig. \ref{fig_7_days}, respectively. The visualized results indicate that the proposed method can generate robust and reliable results for SST forecasting.

\begin{figure}[htb]
  \centering
  \includegraphics[width=3.0in]{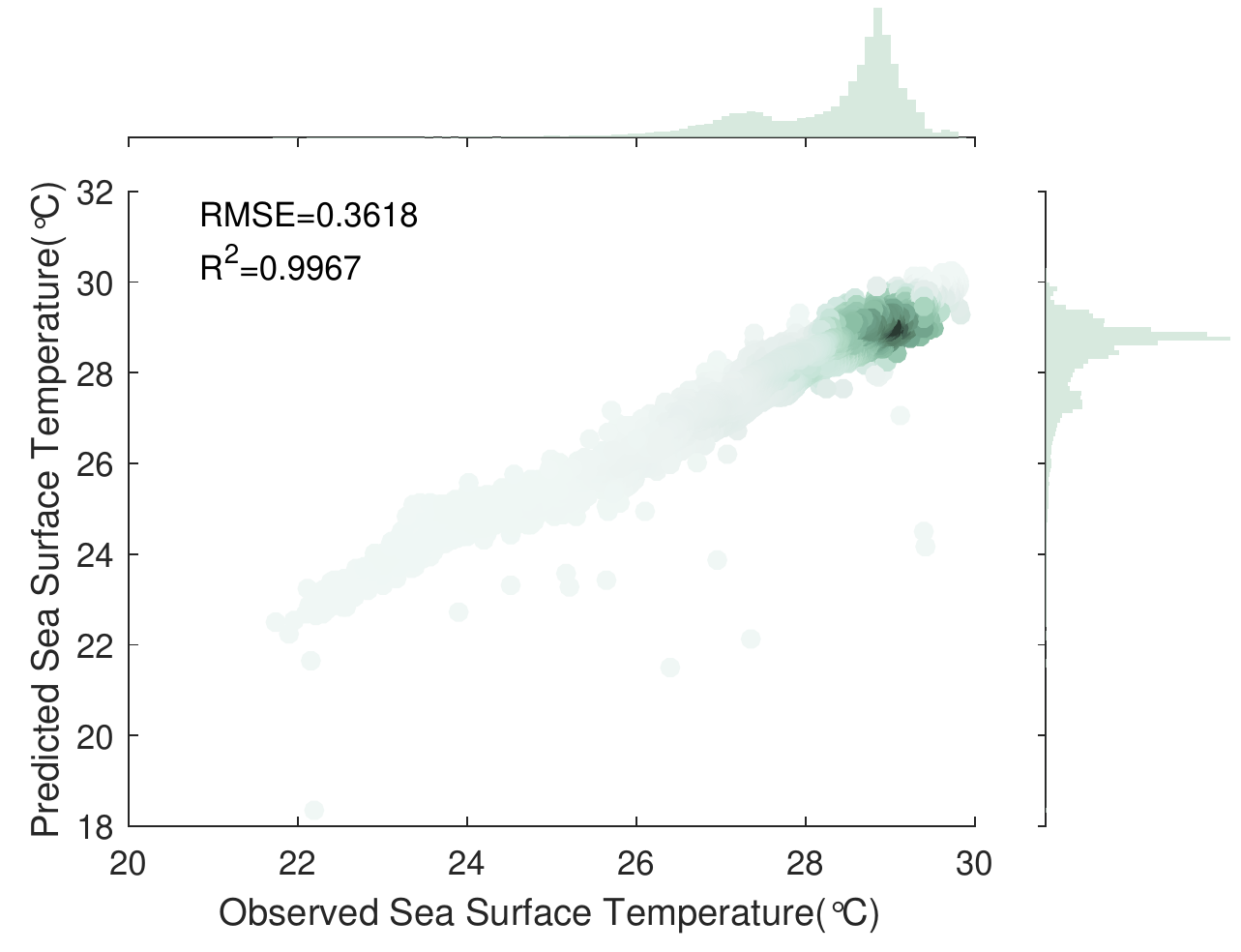}
  \caption{Scatter plot comparing the next one-day SST prediction with the corresponding observed data}
  \label{fig_scatter_1}
\end{figure}

\begin{figure}[htb]
  \centering
  \includegraphics[width=3.0in]{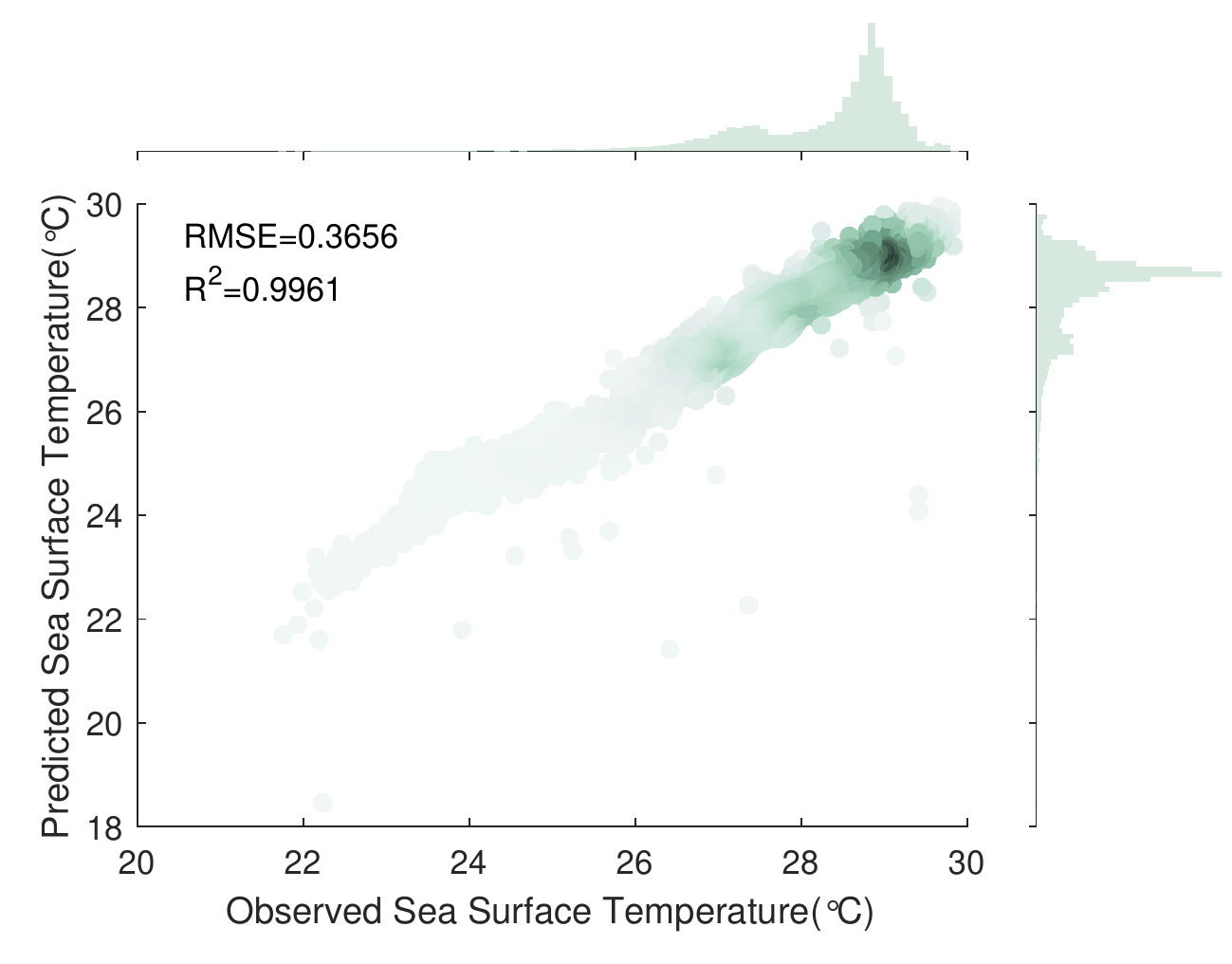}
  \caption{Scatter plot comparing the next three-day SST prediction with the corresponding observed data}
  \label{fig_scatter_3}
\end{figure}

\begin{figure}[htb]
  \centering
  \includegraphics[width=3.0in]{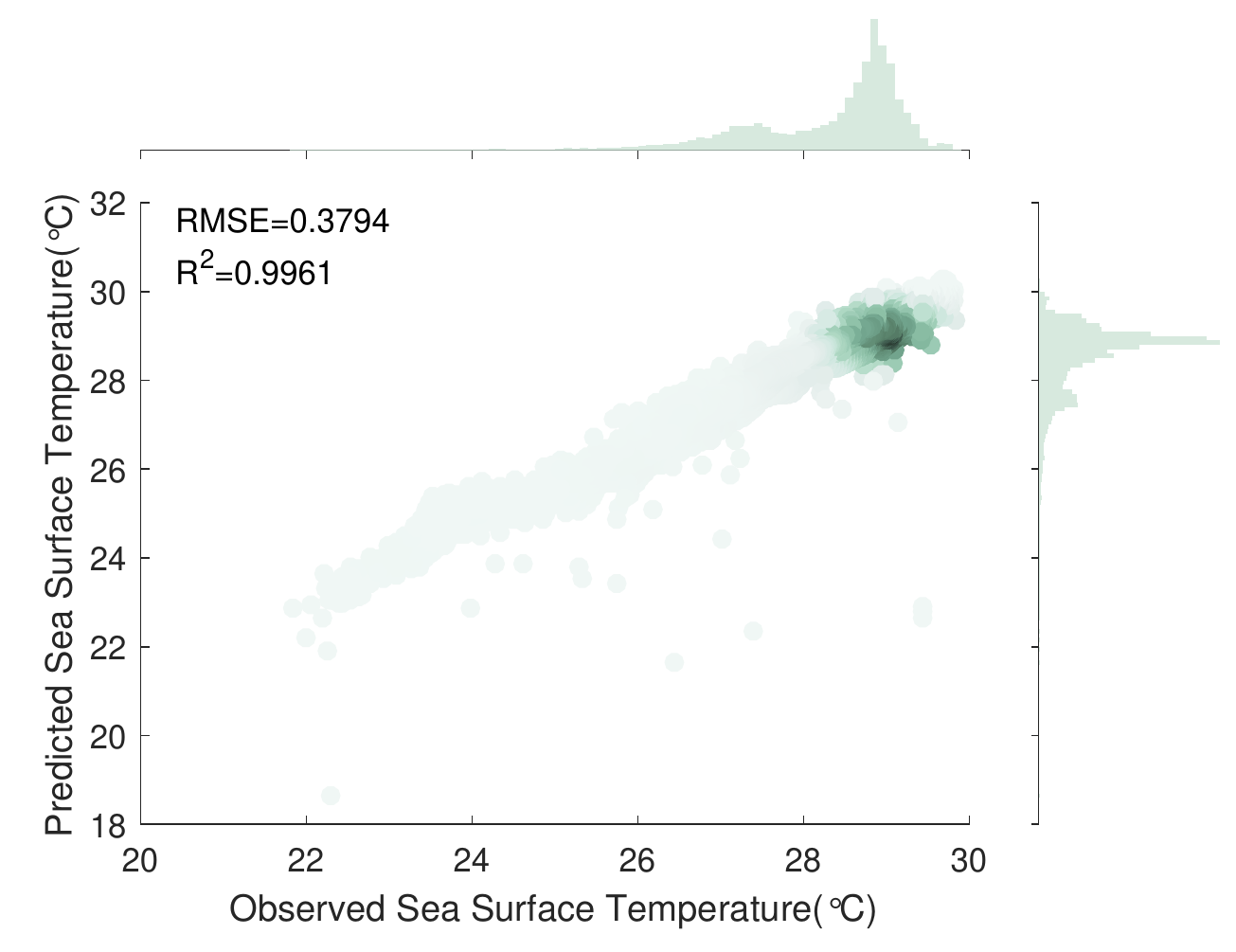}
  \caption{Scatter plot comparing the next seven-day SST prediction with the corresponding observed data}
  \label{fig_scatter_7}
\end{figure}

\begin{figure*}[ht]
\begin{center}
\includegraphics[width=0.8\textwidth]{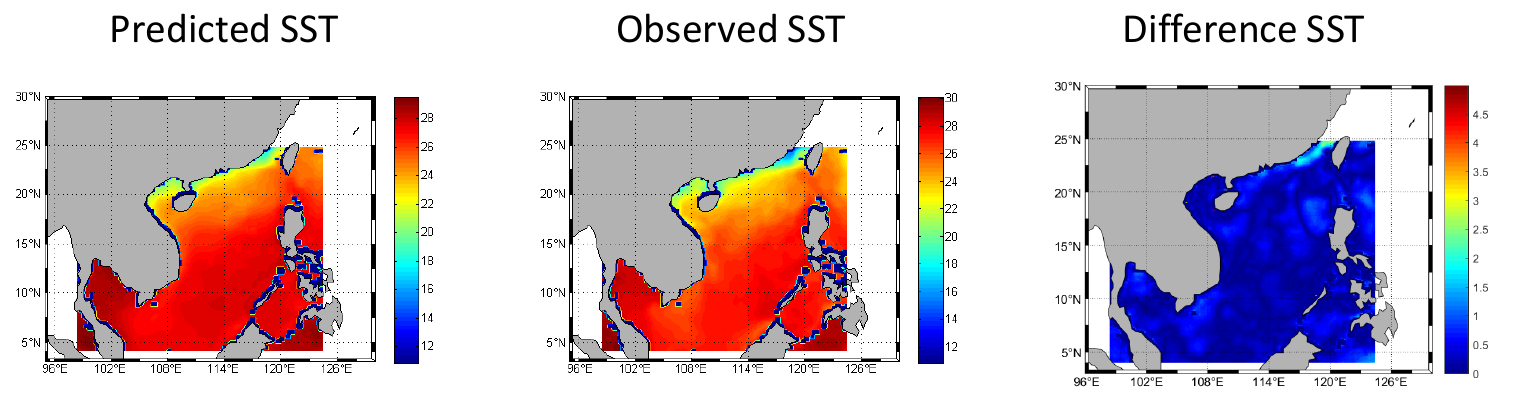}
\caption{Visualized results for the next one-day SST prediction.}
\label{visual_res_1}
\end{center}
\end{figure*}

\begin{figure*}[ht]
\begin{center}
\includegraphics[width=0.8\textwidth]{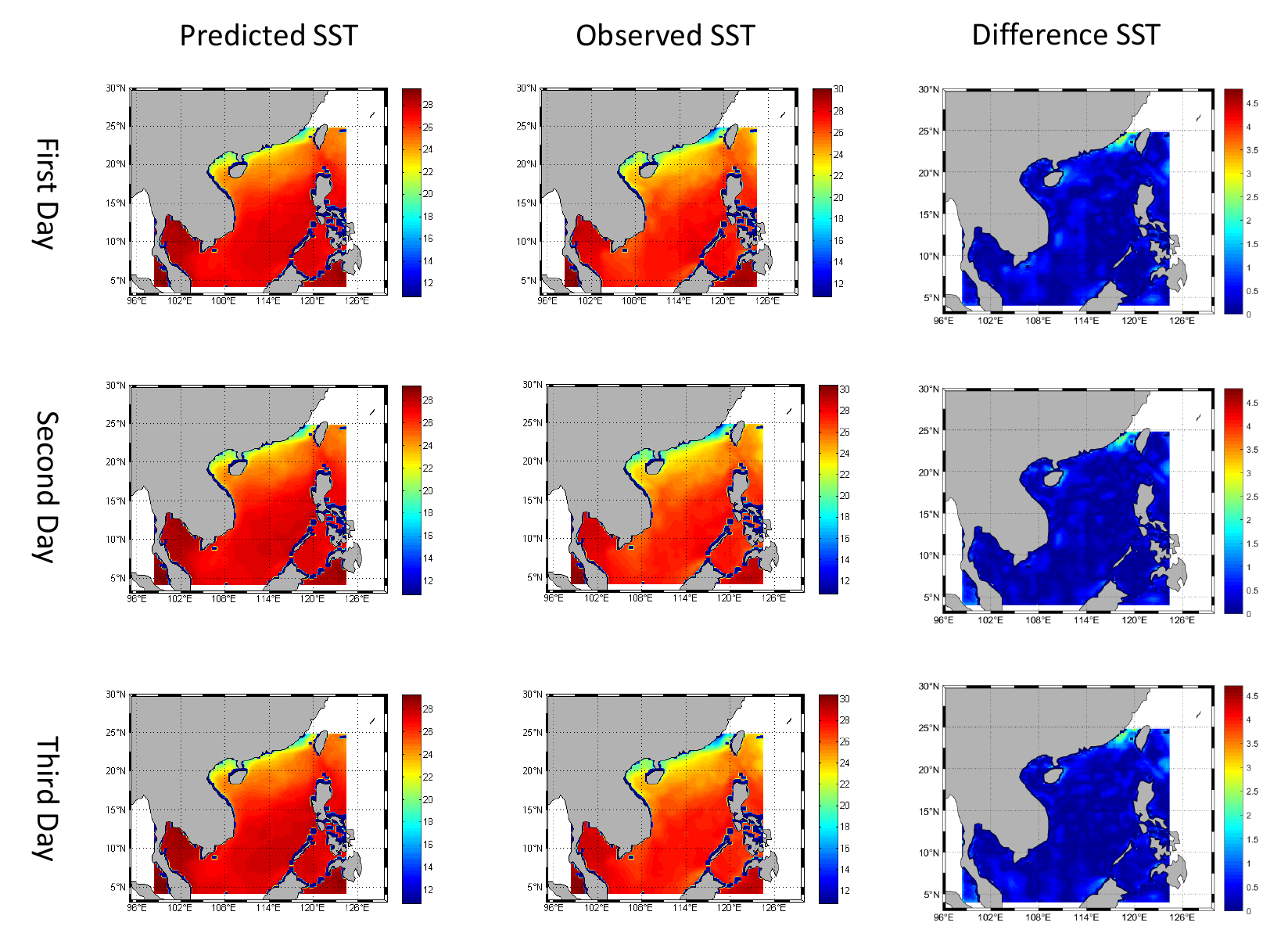}
\caption{Visualized results for the next three-day SST prediction. The first column shows the results of predicted SST. The ground truth observed SST data are shown in the second column. We present their difference in the third column.}
\label{visual_res_2}
\end{center}
\end{figure*}

\begin{figure*}[htbp]
\begin{center}
\includegraphics[width=0.8\textwidth]{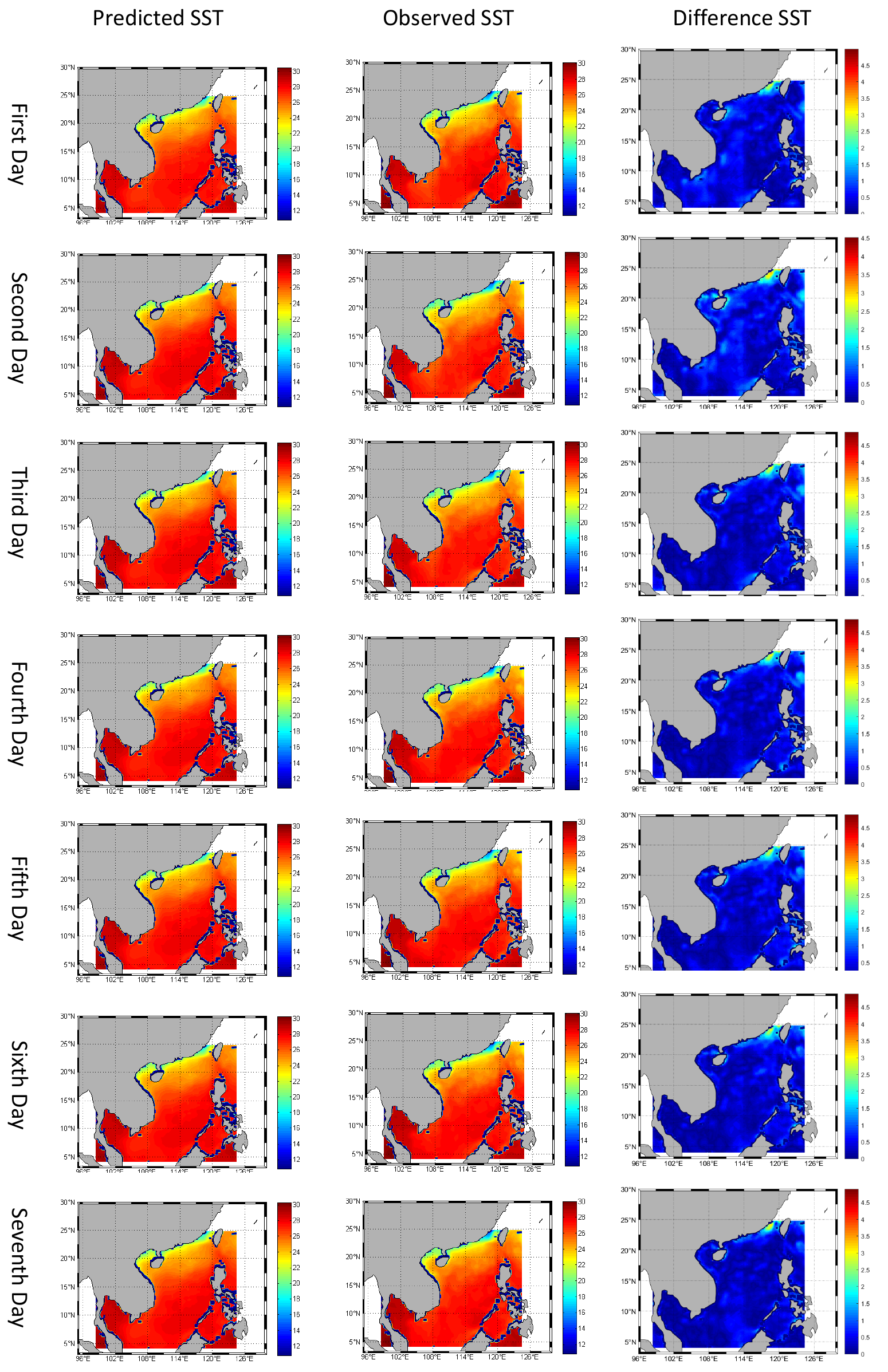}
\caption{Visualized results for the next seven-day SST prediction. The first row shows the results of predicted SST. The ground truth observed SST data are shown in the second row. We present their difference in the third row.}
\label{visual_res_3}
\end{center}
\end{figure*}

A scatter plot on the SST prediction for the next one day is illustrated in Fig. \ref{fig_scatter_1}. It can be observed that the data points are roughly evenly distributed near the red line. Fig. \ref{fig_scatter_3} and Fig. \ref{fig_scatter_7} are scatter plots of the prediction results for the next three days and the next seven days, respectively. The scatter plots demonstrate the effectiveness of the proposed method for SST prediction.

\begin{table}[bp!]
\caption{SST Prediction Results of Different Methods} \renewcommand\arraystretch{1.5}
\label{table_different_methods} 
\begin{center}
\begin{tabular}{|c||c|c|c|c|c|c|}
\hline
\multirow{2}{*}{Model}  & \multicolumn{3}{c|}{RMSE ($^{\circ}$C) $\downarrow$} \\
\cline{2-4} & 1 day & 3 days & 7 days \\ \hline\hline
ConvLSTM & 0.8541$\pm$0.0157& 0.7824$\pm$0.0029 & 0.7712$\pm$0.0215 \\ 
Hybrid-NN &0.9582$\pm$0.0041 &0.9667$\pm$0.0021 &0.9719$\pm$0.0047\\
Hybrid-TL & 0.7786$\pm$0.0024 & 0.8249$\pm$0.0033 & 0.8289$\pm$0.0131\\ 
Gen-END &0.9623$\pm$0.0021 &0.9178$\pm$0.0039 &0.8421$\pm$0.0208\\
VAE-GAN &0.5571$\pm$0.0025 &0.5650$\pm$0.0045 &0.5508$\pm$0.0055\\
Tra-NM  & 1.6134$\pm$0.0031 & 1.6158$\pm$0.0037 & 1.6289$\pm$0.0052 \\
Tra-ASL &0.6681$\pm$0.0014 &0.6829$\pm$0.0165 &0.7166$\pm$0.0374\\
Ours~~~ & \textbf{0.3618$\pm$0.0017}  & \textbf{0.3656$\pm$0.0015} & \textbf{0.3794$\pm$0.0014} \\ \hline \hline
\multirow{2}{*}{Model}  & \multicolumn{3}{c|}{$R^{2}$ $\uparrow$} \\
\cline{2-4} & 1 day & 3 days & 7 days \\ \hline   \hline
ConvLSTM & 0.9729$\pm$0.0073 & 0.9721$\pm$0.0003 & 0.9685$\pm$0.0006 \\ 
Hybrid-NN &0.9917$\pm$0.0005 &0.9917$\pm$0.0002 &0.9914$\pm$0.0003\\
Hybrid-TL & 0.9671$\pm$0.0019 & 0.9663$\pm$0.0008 & 0.9690$\pm$0.0031 \\ 
Gen-END &0.9084$\pm$0.0013 &0.9081$\pm$0.0021 &0.9065$\pm$0.0029\\
VAE-GAN &0.9843$\pm$0.0023 &0.9834$\pm$0.0018 &0.9820$\pm$0.0017\\
Tra-NM  & 0.9182$\pm$0.0016 & 0.9234$\pm$0.0023 & 0.9179$\pm$0.0031 \\
Tra-ASL &0.9603$\pm$0.0016 &0.9593$\pm$0.0016 &0.9582$\pm$0.0005\\
Ours~~~ & \textbf{0.9967$\pm$0.0012} & \textbf{0.9961$\pm$0.0013} & \textbf{0.9961$\pm$0.0024} \\ \hline
\end{tabular}
\end{center}
\vspace*{-2mm}
\end{table}

In order to verify the effectiveness of the proposed method, we compare the proposed method with seven closely related methods: ConvLSTM \cite{shix}, Hybrid-NN \cite{patil_k}, Hybrid-TL \cite{ham_nature}, Gen-END \cite{traditional_encoder}, VAE-GAN \cite{vae-gan}, Tra-NM, and Tra-ASL. The study area is (\ang{3.99}N$\sim$\ang{24.78}N, \ang{98.4}E$\sim$ \ang{124.4}E) for these methods. All of these methods used the training data from the past 5 days for the next 1-day prediction, the data from the past 7 days for the next 3-day prediction, and the data from the past 10 days for the next 7-day prediction.

ConvLSTM is discussed in Section III. C, and it is an effective spatial-temporal model for SST prediction. Hybrid-NN utilizes the discrepancy between the observed data and the numerical model data to guide the training of deep neural networks. Hybrid-TL combines the advantages of numerical models and neural networks through transfer learning. Gen-END is a generative encoder that can be used for SST prediction. VAE-GAN integrates variational autoencoder and GAN, and it can capture high-level semantic features for SST prediction. HYCOM SST data are used to train the ConvLSTM model for the next 1-day, 3-day, and 7-day prediction (termed as Tra-NM). Tra-ASL is a traditional assimilation method and it exploits the correlations among multiple types of data (observed data and numerical model data).

The GHRSST data is first utilized to train a ConvLSTM model, which serves as the baseline. It is a widely used data-driven approach for SST prediction. Hybrid-NN, Hybrid-TL, Gen-END, and VAE-GAN employ the GHRSST and HYCOM data for training. The HYCOM assimilation data \cite{hycom_data} are used here, with a spatial resolution of 1/\ang{12}$\times$1/\ang{12}. Our method improves and rectifies the incorrect components in the numerical model data by introducing physical knowledge from the historical observed data. The corrected numerical model data is referred to as the physics-enhanced data. To compare with the physics-enhanced data, HYCOM assimilation data (Tra-ASL) and HYCOM data (Tra-NM) are similarly used to train the ConvLSTM model.

The training times of ConvLSTM, Hybrid-NN, Tra-NM, and Tra-ASL for the next 1-day, 3-day, and 7-day predictions are 1.8, 4.4, and 8.2 hours, respectively. The Hybrid-TL method trained the ConvLSTM model twice, and the training duration is 3.6, 8.8, and 16.4 hours for the three tasks, respectively. The VAE-GAN requires 181.6, 184.2, and 188.4 hours for training, while the Gen-END method requires almost the same amount of time, with 196.8, 199.3, and 203.2 hours for three SST prediction tasks, respectively.

The results for the next 1-day, 3-day, and 7-day SST prediction are presented in Table \ref{table_different_methods}. It is evident that the Tra-NM method yields unsatisfactory results compared to the other methods. This is likely due to the incorrect components in the HYCOM data, which adversely affect the SST prediction performance. The Hybrid-NN method also performs poorly, as its average RMSE values are the second lowest among the models. The Hybrid-TL model performs better than the ConvLSTM for the next 1-day SST prediction, but not for the other two tasks. Our method achieves the best RMSE values and highest $R^{2}$ values. Compared to the ConvLSTM model, the average RMSE values of our method are effectively improved. It demonstrates that introducing physical knowledge from the observed data can restore the incorrect components in the numerical model data, thus improving the SST prediction accuracy.

Fig. \ref{visual_res_1} presents the visualized results for the next one-day SST prediction, the observed SST data, and their differences, respectively. It can be seen that the predicted results are highly similar to the observed SST data across the entire region of the South China Sea. Fig. \ref{visual_res_2} displays the visualized results for the next three-day SST prediction. It is observed that there are some significant difference values in the Gulf of Tonkin and in other marginal areas of the South China Sea. Fig. \ref{visual_res_3} illustrates the visualized results for the next seven-day SST prediction. It is found that the major difference values mainly concentrate on the Gulf of Tonkin for the next seven-day prediction, and they are larger than the results for the two other tasks.

\subsection{Limitation and Discussion}

\begin{figure}[htb]
  \centering
  \includegraphics[width=2.2in]{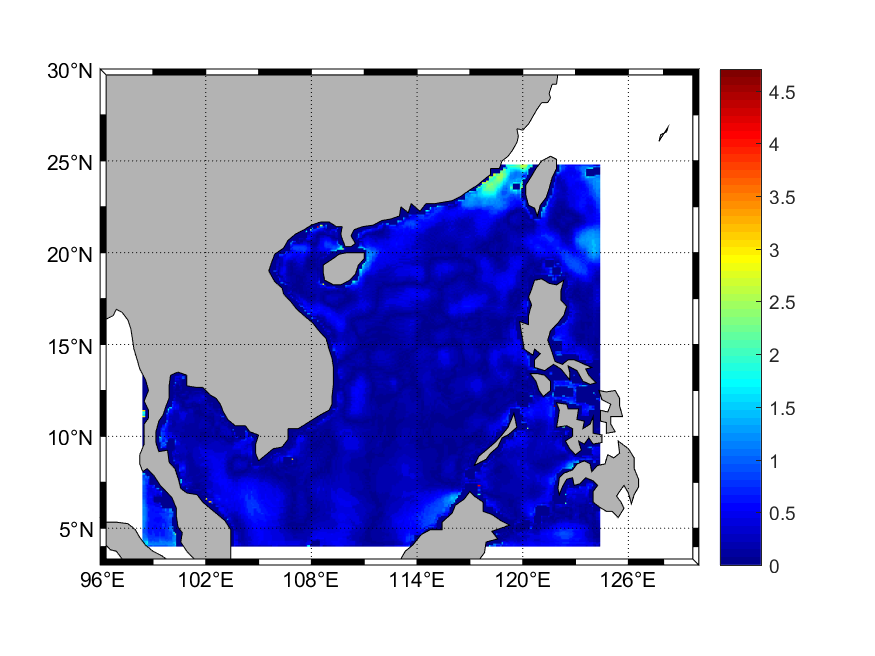}
  \caption{Visualization of the SST prediction error.}
  \label{fig_vis_error}
\end{figure}

From Fig. 7 to Fig. 9, it can be observed that there are some inaccuracies in the mid-range SST, which are visualized in Fig. \ref{fig_vis_error}. Bright pixels indicate large SST prediction errors, whereas dark pixels denote accurate SST predictions. As can be seen, these points are mainly located on the northwestern part of Taiwan Strait, where the predicted sea surface temperature is lower than the observed data.  The prediction error is mainly caused by the ConvLSTM model and the land mask. In our implementations, the land mask is applied to the study area. The ConvLSTM exploits the spatial and temporal features of the whole study area. The features of the northwestern part of Taiwan Strait are affected by the land mask to some extent, and therefore result in prediction errors. If higher resolution training data could be obtained, the accuracy of predictions in this region would be further improved.

In Fig. 11 and 12, it can be seen that there is no significant increase in errors with the lead day. This may be due to the fact that our method uses a sufficient amount of training data, and the deep neural networks are able to effectively capture the temporal features. Furthermore, the persistence of SST is also an important factor.

\section{Conclusions and Future Work}

In this paper, we present a SST prediction approach based on physical knowledge correction, which utilizes historical observed data to refine and adjust the physical component in the numerical model data. Specifically, a prior network was employed to extract physical knowledge from the observed data. Subsequently, we generated physics-enhanced SST by applying the pretrained prior network over numerical model data. Finally, the generated data were used to train the ConvLSTM network for SST prediction. Additionally, the physical knowledge-based enhanced data were leveraged to train the ConvLSTM network, which further improved the prediction performance. The proposed method achieved the best performance compared to six state-of-the-art methods. 

Although the physical part of the numerical model data has been corrected by our proposed method, the prediction performance could be further improved if an interpretable model is employed. In the future, we plan to extract more pertinent knowledge from the deep networks, and then design interpretable models more suitable for practical applications.

\begin{IEEEbiography}[{\includegraphics[width=1in,height=1.25in,clip,keepaspectratio]{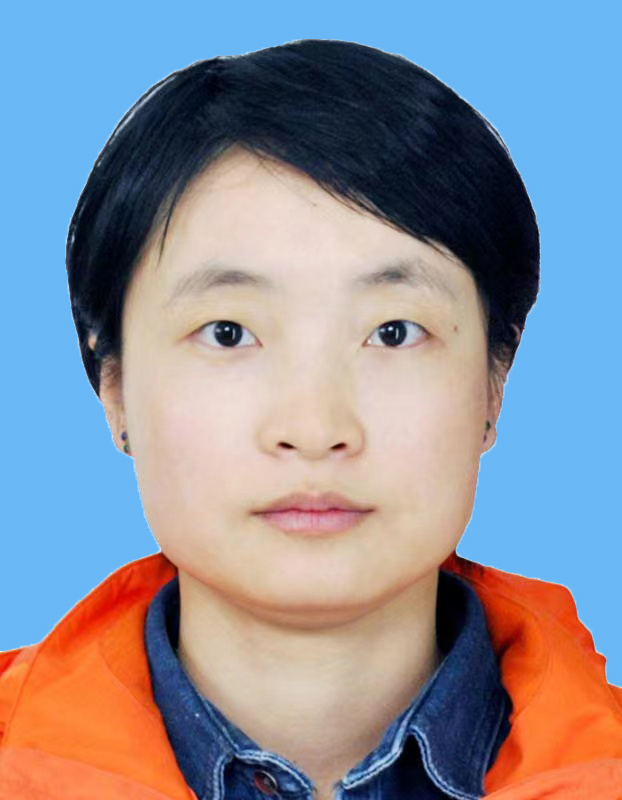}}]{Yuxin Meng} 
received the B.Eng. degree in computer science and technology from the Anhui University of Science and Technology, Huainan, China, in 2010. She is currently pursuing the Ph.D. degree with the Vision Lab, Ocean
University of China, Qingdao, China, supervised by
Prof. Junyu Dong.

Her research interests include image processing
and computer vision.
\end{IEEEbiography}

\begin{IEEEbiography}[{\includegraphics[width=1in,height=1.25in,clip,keepaspectratio]{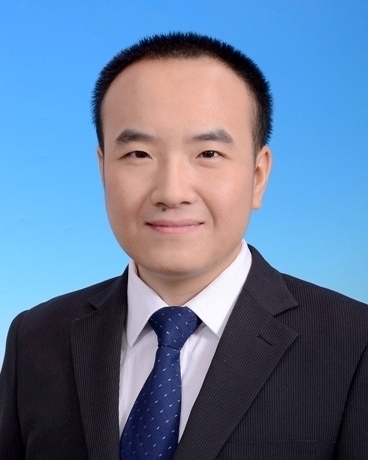}}]{Feng Gao} (Member, IEEE)
received the B.Sc degree in software engineering from Chongqing University, Chongqing, China, in 2008, and the Ph.D. degree in computer science and technology from Beihang University, Beijing, China, in 2015.

He is currently an Associate Professor with the School of Information Science and Engineering, Ocean University of China. His research interests include remote sensing image analysis, pattern recognition and machine learning.
\end{IEEEbiography}

\begin{IEEEbiography}[{\includegraphics[width=1in,height=1.25in,clip,keepaspectratio]{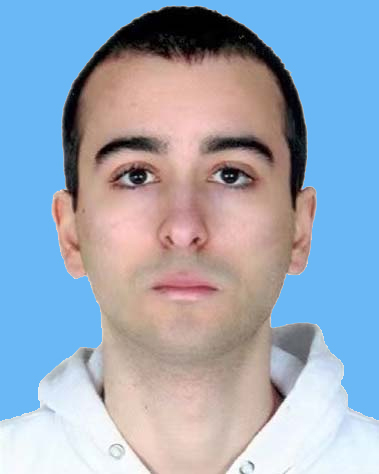}}]{Eric Rigall}
received the Engineering degree from the Graduate School of Engineering, University of Nantes, Nantes, France, in 2018. He is currently pursuing the Ph.D. degree with the Vision Laboratory, Ocean University of China, Qingdao, China, supervised by Prof. Junyu Dong. 

His research interests include radio-frequency identification (RFID)-based positioning, signal and image processing, machine learning, and computer vision.

\end{IEEEbiography}

\begin{IEEEbiography}[{\includegraphics[width=1in,height=1.25in,clip,keepaspectratio]{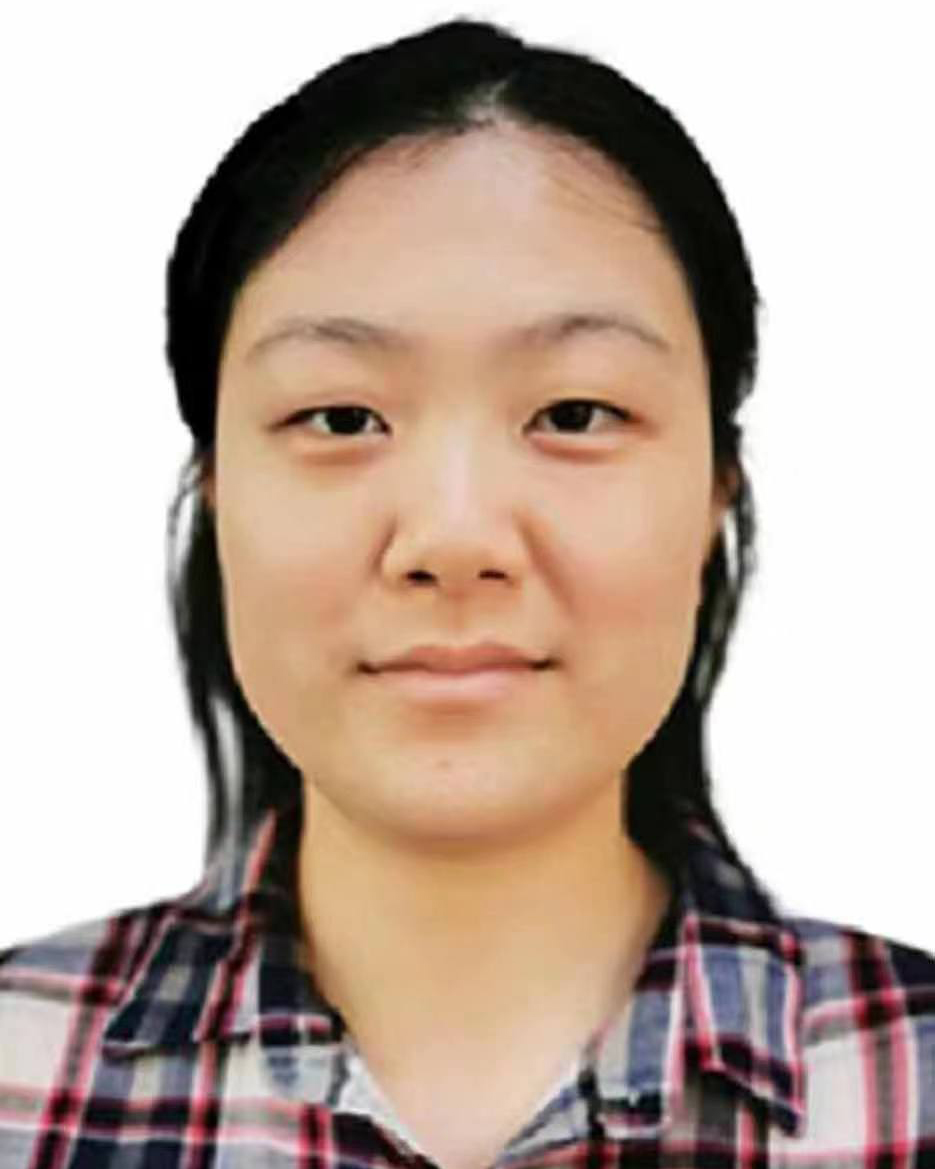}}]{Ran Dong}
received the B.Sc degree in Mathematics and Statistics from Donghua University, Shanghai, China, in 2014, and the Ph.D. degree in Mathematics and Statistics from University of Strathclyde, United Kingdom, in 2020.

She is currently a Lecturer with the School of Mathematical Science, Ocean University of China. Her research interests include artificial intelligence, mathematics, and statistics.
\end{IEEEbiography}

\begin{IEEEbiography}[{\includegraphics[width=1in,height=1.25in,clip,keepaspectratio]{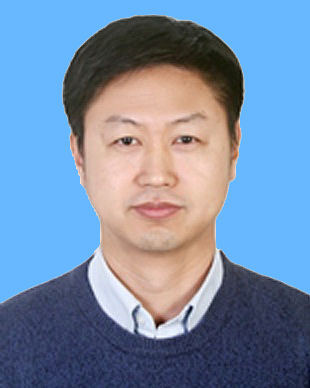}}]{Junyu Dong}
 (Member, IEEE) received the B.Sc. and M.Sc. degrees from the Department of Applied Mathematics, Ocean University of China, Qingdao, China, in 1993 and 1999, respectively, and the Ph.D. degree in image processing from the Department of Computer Science, Heriot-Watt University, Edinburgh, United Kingdom, in 2003.

He is currently a Professor and Dean with the School of Computer Science and Technology, Ocean University of China. His research interests include visual information analysis and understanding, machine learning and underwater image processing.
\end{IEEEbiography}

\begin{IEEEbiography}[{\includegraphics[width=1in,height=1.25in,clip,keepaspectratio]{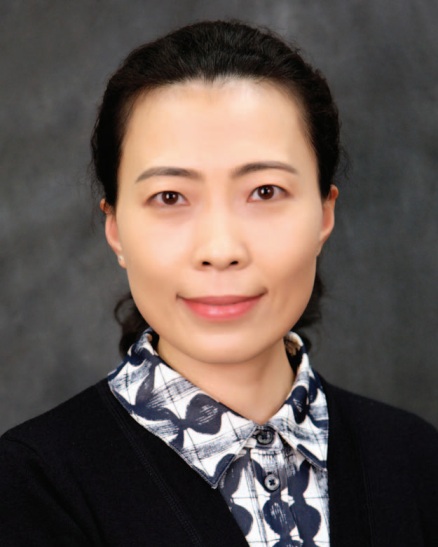}}]{Qian Du}
(Fellow, IEEE) received the Ph.D. degree in electrical engineering from the University of Maryland at Baltimore, Baltimore, MD, USA, in 2000.

She is currently the Bobby Shackouls Professor with the Department of Electrical and Computer Engineering, Mississippi State University, Starkville, MS, USA. Her research interests include hyperspectral remote sensing image analysis and applications, and machine learning.

Dr. Du was the recipient of the 2010 Best Reviewer Award from the IEEE Geoscience and Remote Sensing Society (GRSS). She was a Co-Chair for the Data Fusion Technical Committee of the IEEE GRSS from 2009 to 2013, the Chair for the Remote Sensing and Mapping Technical Committee of International Association for Pattern Recognition from 2010 to 2014, and the General Chair for the Fourth IEEE GRSS Workshop on Hyperspectral Image and Signal Processing: Evolution in Remote Sensing held at Shanghai, China, in 2012. She was an Associate Editor
for the \textsc{Pattern Recognition}, and \textsc{IEEE Transactions on Geoscience and Remote Sensing}. From
2016 to 2020, she was the Editor-in-Chief of the \textsc{IEEE Journal of Selected Topics in Applied Earth Observation and Remote Sensing}. She is currently a member of the IEEE Periodicals Review and Advisory Committee and SPIE Publications Committee. She is a Fellow of SPIE-International Society for Optics and Photonics (SPIE).

\end{IEEEbiography}

\end{document}